\documentclass{ceurart}

\usepackage{amsmath, amssymb, amscd, amsthm, amsfonts}
\usepackage{listings}
\usepackage{graphicx}
\usepackage{subcaption}
\usepackage{caption}
\begin{document}

\copyrightyear{2024}
\copyrightclause{Copyright for this paper by its authors.
  Use permitted under Creative Commons License Attribution 4.0
  International (CC BY 4.0).}

\author[1]{Sumit Paul}[%
orcid=0009-0007-4110-9183,
email=sumit.paul@tu-berlin.de,
]
\author[2]{ Dr. Danh Lephuoc}[%
email=danh.lephuoc@tu-berlin.de,
]

\author[3]{ Prof. Dr. Manfred Hauswirth}[%
email=manfred.hauswirth@tu-berlin.de,
]

\title{Performance Evaluation of ROS2-DDS middleware implementations facilitating Cooperative Driving in Autonomous Vehicle.}

\copyrightyear{2024}

\copyrightclause{Copyright for this paper by its authors. Use permitted under Creative Commons License Attribution 4.0 International (CC BY 4.0).}

\conference{Edge AI meets Swarm Intelligence Technical Workshop, September 18, 2024, Dubrovnik, Croatia}

\begin{abstract}
In the autonomous vehicle and self-driving paradigm, cooperative perception or exchanging sensor information among vehicles over wireless communication has added a new dimension. Generally, an autonomous vehicle is a special type of robot that requires real-time, highly reliable sensor inputs due to functional safety. Autonomous vehicles are equipped with a considerable number of sensors to provide different required sensor data to make the driving decision and share with other surrounding vehicles. The inclusion of Data Distribution Service(DDS) as a communication middleware in ROS2 has proved its potential capability to be a reliable real-time distributed system. In order to provide an abstraction to the ROS2 users and to utilize DDS functionality, ROS2 messages are converted into the DDS format by ROS2 for transmission. DDS comes with a scoping mechanism known as domain. Whenever a ROS2 process is initiated, it creates a DDS participant. It is important to note that there is a limit to the number of participants allowed in a single domain.

The efficient handling of numerous in-vehicle sensors and their messages demands the use of multiple ROS2 nodes in a single vehicle. Additionally, in the cooperative perception paradigm, a significant number of ROS2 nodes can be required when a vehicle functions as a single ROS2 node. These ROS2 nodes cannot be part of a single domain due to DDS participant limitation; thus, different domain communication is unavoidable. Moreover, there are different vendor-specific implementations of DDS, and each vendor has their configurations, which is an inevitable communication catalyst between the ROS2 nodes. The communication between vehicles or robots or ROS2 nodes depends directly on the vendor-specific configuration, data type, data size, and the DDS implementation used as middleware; in our study, we evaluate and investigate the limitations, capabilities, and prospects of the different domain communication for various vendor-specific DDS implementations for diverse sensor data type.

\end{abstract}



\maketitle

\section{Introduction}

Autonomous vehicles are a specialized version of robots that have to perform a variety of maneuvers such as driving at high speed on highways, parking precisely, overtaking or lane changing, and avoiding obstacles. All of these scenarios should be handled by the autonomous vehicle with high functional safety because it has direct interaction with the people who are traveling, walking on the walkway, or riding on the way back home. So the deployed computer systems in the autonomous vehicle have to process lots of sensor information to decide the correct driving decisions\cite{Helper1.1}. Cooperative perception provides extended sensing capabilities beyond line-of-sight and field-of-view through the exchange of sensor, camera, or radio device data. This extra perception range provided by the cooperative perception in the field of autonomous vehicles can eventually improve the safety of the traffic and improve the efficiency of driving decisions.\cite{Cooperative0.4} \cite{Cooperative0.5}. Authors of \cite{Helper1.1} have proposed and developed an autonomous vehicle architecture with Robot Operating System 2. The reliability and real-time performance of the second-generation Robot Operating System \cite{ROS2}  made it suitable for autonomous vehicles, as it supports the fundamental publish-subscribe messaging mechanism like ROS1 \cite{ROS1} and a dedicated, scalable, fault-tolerant, reliable, secure as well as durable middleware like DDS(Data Distribution System)\cite{DDS}. 

The initial version of the Robot Operating System (ROS1) had several shortcomings, including operating system dependency, lack of fault tolerance, process synchronization issues, and time constraints, which clearly disqualified it from being a real-time system. The Data Distribution Service (DDS) standard by the Object Management Group (OMG) introduced a robust publish/subscribe model with Quality of Service (QoS) for dependable and timely data propagation. DDS supports various transport configurations, ensuring safety, scalability, resilience, security, and fault tolerance, making it the perfect fit for distributed systems. It effectively utilizes a Data-Centric Publish-Subscribe (DCPS) model with components such as topics, data readers, data writers, publishers, and subscribers. The DDS domain, identified by a domain ID, enables seamless communication between publishers and subscribers through topics within the Global Data Space (GDS), thereby ensuring complete isolation.

Even though the GDS(Global Data Space) and the scoping mechanism of the domain make the publish/subscribe communication easier, it brings some concerns when we have a specific number of domain participants and topics. A DDS participant is created when a ROS2 processes spin up on a machine. This single participant utilizes two ports on the system. So more than 120 ROS2 processes on a single machine cannot reside in the same domain. Moreover, for managing a considerable number of sensors and exchanging their generated data, a single vehicle or robot requires multiple ROS2 nodes \cite{Helper1.1}.  Regarding cooperative perception for automated vehicles, the vehicles or robots act as ROS2 nodes and exchange data among them over wireless communication \cite{CooperativeDriving}. In this case of a cooperative perception for automated vehicles, it is easily possible to have an unlimited number of ROS2 nodes communicating with each other. Additionally, in one of the vendor-specific implementations of DDS, RTI-Connext-DDS for version 4.1e and lower, only 58 participants, and for version 4.2e and higher, 120 participants are allowed to join in one domain \cite{RTIMaxDomainParticipants}. On the other hand, as the DataReader and DataWriter of DDS are directly associated with one topic so the number of domain participants indirectly determines the number of topics that can be created in one domain \cite{RTIMaxTopics}. As mentioned before, in the autonomous and cooperative perception paradigm, large amounts of sensor data must be transferred among different connected vehicles. It is improbable to use only a single domain with a limited number of participants and topics. To establish the communication context for a higher number of participants and to fulfill the requirement of a flexible number of topics, the usages of multiple domains are unavoidable. DDS leverages UDP/IP multicast as one of the very core foundations, so it is difficult to scale the communication over multiple LANs or a WAN \cite{Zenoh}. Different vendor-specific DDS implementations have their specific way of establishing communication for rapid scaling and integrating various nodes running in disparate domains or geographically dispersed. These linking applications work as a bridge across the different domains to establish a system-of-systems architecture.

\begin{table}[h!]
  \begin{center}
    \caption{Vendor-specific DDS implementations and bridging services.}
    \label{tab:ddslist}
    \begin{tabular}{|l|p{2in}|p{2in}|} 
    \hline
      \textbf{DDS Implementation} & \textbf{Bridging Application} & \textbf{Configuration}\\
      \hline
        Eclipse Cyclone DDS &  Zenoh-plugin-DDS & Publisher and Subscriber both require to have zenoh shared library \\
     \hline
     eProsima Fast-DDS & Integration Service  & Configuration is done with the YAML file.\\
      \hline
        RTI-Connext DDS & Routing Service & Configuration is done with XML file.\\
      \hline
  
    \end{tabular}
  \end{center}
\end{table}

Each bridging service has its own configuration, dependent libraries, and building process. The communication between two ROS2 nodes heavily depends on various factors, such as the DDS implementation chosen from a specific vendor, configuration settings, the type and size of data transmitted, and the communication medium employed. It is cumbersome for a robotics application developer to test every single possible bridging application before designing the whole echo system. There is also no guarantee of which service reacts how to latency characteristics with different types of sensor data. It is also inconceivable that in the real world, all the robots use the same ROS2 versions with similar underlying DDS versions. According to our research, we have not found any performance evaluation which depicts the performance of vendor-specific DDS implementations in the case of different domain communication.
\\
\\
\textbf{ Contribution}\\
Our research has found no performance evaluations demonstrating how vendor-specific DDS implementations perform in different domain communication scenarios for different physical machines and data types. There are also no performance evaluations available for the newly released version ROS2-Humble. This paper presents a noble performance comparison between the same domain and different domain communication among ROS2 nodes. The results of our study will assist developers and researchers in understanding how communication characteristics shift when exchanging data between robots in the same domain versus different domains. Our experiments involve using real-life physical devices such as Raspberry Pi3, Raspberry Pi4, Jetson Nano, and laptop computers, enabling robot system developers to understand how product requirements correlate with the device type. By utilizing different data types and communication mediums in our evaluation process, we aim to provide insight into which data should be used and which communication medium should be chosen for optimal performance.

\section{Related Work}

Research in cooperative driving systems and ROS performance has seen significant advancements, addressing crucial aspects of autonomous vehicle communication and control. A pioneering study by \cite{CooperativeDriving} introduced a cooperative driving system architecture leveraging extended perception beyond line-of-sight, which proved instrumental in assisting with complex maneuvers like overtaking and lane-changing. This research underscored the critical role of wireless communication in enabling cooperative perception, setting the stage for deeper investigations into robotic operating systems. In another research \cite{Kim2013CooperativePF}, the author has exclaimed that the extension of the boundary of the sensing area is possible while the vehicles are networked. Depending on the network connectivity, beyond-line-of-sight sensing is possible for the autonomous vehicle. \\
The performance characteristics of ROS1 and ROS2 became a focal point for researchers seeking to optimize these systems for autonomous vehicles. A comprehensive study by \cite{ROS2.1} provided the first in-depth comparison between ROS1 and ROS2, analyzing key metrics such as latency, memory consumption, and throughput. This research utilized various message sizes and DDS middleware implementations, revealing important trade-offs between Quality of Service (QoS) policies and latencies. Their findings suggested that message fragmentation at 64 KB could optimize performance, a insight that has influenced subsequent research and development.\\
Building on this foundation, \cite{ROS2.2} dived deeper into ROS2's scalability, examining how the number of nodes impacts system performance in a data processing pipeline. This study simulated real-world scenarios by creating a chain of nodes for sensor data publishing, processing, and subscribing, providing valuable insights into ROS2's behavior under varying loads and frequencies.\\
The applicability of ROS2 in real-time systems, crucial for autonomous vehicles, was explored by several researchers. Notably, \cite{Helper1.1} proposed a self-driving car architecture using ROS2, demonstrating that a real-time Linux kernel outperformed standard kernels in high-frequency data publishing scenarios. This finding was corroborated by \cite{Helper1.2}, who conducted extensive tests on ROS2's suitability for real-time robot applications, identifying the Linux Network Stack as a potential bottleneck for achieving hard real-time Ethernet communication. In \cite{Helper1.2}, researchers have done performance analysis among different middlewares like DDS, OPC UA,  MQTT and  ROS1. To measure the round trip time between two nodes, the authors used the request-reply communication pattern, where the server node provided the methods, and the client node called the methods to measure the performance of the server node. They argued over the better performance of OPC UA and DDS over MQTT and ROS1.\\
Security considerations, paramount in autonomous vehicle systems, were thoroughly addressed by \cite{ROS2.3}. Their evaluation of ROS2 performance under various security settings and QoS configurations revealed that security measures have a more significant impact on latency compared to QoS settings. This work highlighted the need for careful consideration of security implementations in time-critical systems. Further exploration of the security-performance trade-off was conducted by \cite{ROS2.7}, who compared different encryption scenarios including no security, cryptographic algorithm-based security (SROS2), and SSL/TLS-based security (OpenVPN). Their findings provided valuable guidelines for selecting appropriate security measures based on specific use case requirements.\\
To enhance ROS2's inter-process communication efficiency, \cite{ROS2.5} introduced the innovative Toward Zero-Copy (TZC) framework. This approach, which divides messages into control and data parts and utilizes shared memory for data transmission, showed remarkable performance improvements, especially for large messages and multiple subscribers. However, its limitation in handling ROS2 strings highlighted areas for further development.
The dynamic nature of autonomous vehicle environments inspired \cite{ROS2.4} to develop a mechanism for dynamically binding multiple DDS implementations in ROS2. This adaptive approach allows systems to optimize communication based on changing environmental conditions and requirements.

\section{Methodology}
In this section, we introduce our parameter scope used in the experiments, which is followed by the measurement metric definition.
\subsection{Parameter Space}\label{parameter}
Our research and experiment parameters originated from our use case of exchanging sensor data between multiple different-domain ROS2 nodes. We assume the sensors produce the sensor data and publish it immediately; as different sensors have different sampling frequencies, we consider variable publisher frequency as one of our most essential parameters. A ROS2 node must provide the specific numeric domain id to participate in a specific domain. As our use case leads toward the communication between different domain ids, we have included multiple domain ids as our parameter. Various types of sensors are equipped with autonomous vehicles, producing different types and sizes of data. Various file types, such as Binary, String, and numeric series, and the variable size of these data types are used as data type and file size parameters, respectively. In order to establish communication between a publisher and subscriber in ROS2, nodes must provide a QoS policy profile. The default setting for the Durability,  Reliability, and History QoS settings are "volatile," "reliable," and "keep last," respectively. The default queue size for History QoS is 10. The Deadline, Liveliness, and Lifespan  QoS settings are set as "system default" according to the default QoS settings \cite{ROS2_default_QoS}. We have used the aforementioned QoS policy settings throughout the sensor data exchange between different domain communications.

\subsection{Measurement Metrics}
\begin{figure}[h]
\centering
  \includegraphics[width=0.8\linewidth]{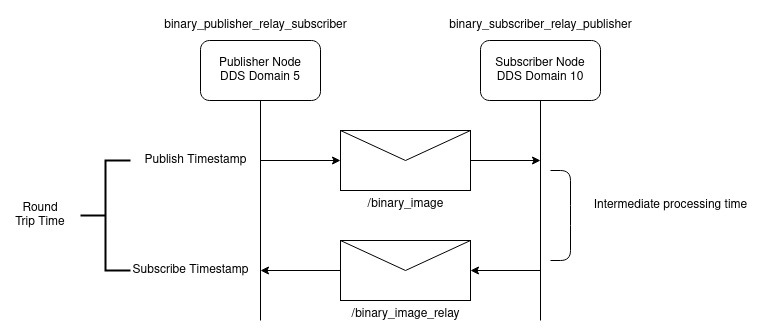}
  \captionsetup{justification=centering}
  \caption{ Round Trip Time.}
  \label{fig:rtt}
\end{figure}

For real-time systems like autonomous vehicles, the latency of the transmitted message plays the most crucial role, so we have considered latency as our primary metric for our measurement metric. In the distributed system architecture, it is challenging to measure one-way latency due to the synchronized clocks in various machines \cite{ROS2.7}. For this reason, we have used Round Trip Time to deduce latency. As our main focus of the evaluation is how the latency characteristic behaves while communication happens through varieties of DDS implementations between different DDS domains, we assume the ROS2 system behaves similarly for the various DDS implementations. \\
\\
\textbf{Round Trip Time} \\
To measure the Round Trip Time(RTT), we have used various physical devices connected via wire and wireless communication mediums as the ROS2 publisher and subscriber node. The total time a ros2 message takes to travel from the publisher node to the subscriber node and from the subscriber node to the publisher node is measured as the Round Trip Time of that message, also known as the ping-pong test figure-\ref{fig:rtt}. We have used custom messages to describe each ROS2 topic because we have different data types of testing. 

For the binary file type, we have used a custom message type BinaryFile which includes the \verb |sensor_msgs/Image| type to hold the binary image and a \verb|float64| data type to hold the publish time. While publishing a frame, a timestamp (T1) is taken and included in the message. The publisher residing in domain 5 creates the topic \verb |binary_image| to publish the binary image file and also creates a subscriber of the topic \verb |binary_image_relay| to receive the replied message from the subscriber node. The ROS2 subscriber node residing in domain 10 creates a subscriber of the topic \verb |binary_image| to receive the message and a publisher component to relay-publish the received message on the topic \verb |binary_image_relay|. When the ROS2 publisher node receives the relay-published message, it takes another timestamp(T2) to get the message received time. We can calculate The Round Trip Time by subtracting T1 from T2. 
 \begin{equation}
      RTT = T2 - T1  
  \end{equation}

For the String and IMU data types, we have used custom file types StringTimestamp and imu, respectively. We followed similar topic patterns as BinaryFile type to define the publisher and relay-subscriber for the String and IMU transfer measurement. \\
\\
\textbf{Latency measurement}\\
The ROS2 subscriber node works as a relay publisher to facilitate the Round Trip Time calculation. The time a ROS2 subscriber node takes to subscribe to a specific topic to receive the data and relay publishes the data is known as intermediate-process time. But while communicating with variable sizes and types of data, the intermediate-processing time for the relay-publisher node also varies accordingly. Which directly affects the Round Trip Time. For example, the intermediate-processing time for 33KB of binary data and  4MB of binary data is different.

To measure the intermediate-process time on the ROS2-subscriber-relay-publisher node, a timestamp (T3) is captured as soon as the subscriber node starts downloading the data frame after subscribing to the desired topic, and another timestamp (T4) is captured when the subscriber node relay-publishes that specific data frame. So the intermediate-process time is measured by subtracting T3 from T4. By doing the intermediate-process calculation on the subscriber node itself, we can avoid the complexity of clock synchronization.
   \begin{equation}
      \operatorname{intermediate-processtime} = T4 -  T3 
  \end{equation}

By subtracting the intermediate-processtime from RTT and dividing it by 2, we can get the latency of a specific data frame. So the latency measurement equation is the following
   \begin{equation}
      Latency = \frac{RTT -  \operatorname{intermediate-processtime}  }{2}
  \end{equation}

\subsection{Experiment Setup}
The experiment setup consists of data, hardware, and software parts. This section presents our experiment setup, which is used to evaluate the performance of different ROS2 DDS implementations from the viewpoint of test data, physical system, and software stack. \\

\noindent\textbf{Data }\\
In this section \ref{parameter}, we have already mentioned the desired parameters for our experiments. In the cooperative autonomous driving paradigm, to detect an object or obstacle or traffic signal, Binary type of data is used, and to share driving decision-related data, String and IMU type data is used. We have collected the primary Binary and IMU type of data from the \href{https://projects.asl.ethz.ch/datasets/doku.php?id=kmavvisualinertialdatasets}{ASL Dataset}. But this data set's binary files or images are of one size, which does not fulfill our requirement of varying file sizes. To resolve this issue, 
we randomly chose different binary files, which are free from the internet, and using our script, we created a data set of various binary file sizes. We have taken different lengthy strings from \href{ https://vedabase.io/en/library/bhakti/1/#:~:text=Lord%20Caitanya%20met%20the%20two,service%20and%20join%20Lord%20Caitanya.}{this book} for the String file type randomly. The IMU data is always a fixed size data, so we used the IMU data from the ASL Dataset. To verify our evaluation framework for the real-life live data, we have run the experiment framework with Jetson Nano \cite{Jetson} connected with a Realsence camera \cite{Realsense}.\\

\noindent\textbf{Hardware }\\
We have used different physical devices, as ROS2 publisher and subscriber node, for the performance evaluation of different DDS implementations. The below-mentioned table \ref{tab:hardware} presents the hardware specification of the physical devices.
\begin{table}[h!]
  \begin{center}
    \caption{Hardware specification of physical devices.}
    \label{tab:hardware}
    \begin{tabular}{|c|c|c|} 
    \hline
      \textbf{Machine} & \textbf{CPU} & \textbf{Memory} \\
      \hline
        PC-1 & 8-core x86\textunderscore64 1.90 GHz   &  16GB SODIMM DDR4 2400 MHz 64bit   \\
     \hline
      PC-2 &  4-core x86\textunderscore64 2.30 GHz &  8GB SODIMM DDR4 2133 MHz 64bit\\
      \hline
       Pi 3 & 4-core ARM-A53 1.20 GHz & 1Gb LPDDR4 900 MHz 32 bit\\
      \hline
      Pi 4 & 4-core ARM-A72 1.50 GHz & 8Gb LPDDR4 3200 MHz 32 bit\\
      \hline
      Jetson Nano & 4-core ARM-A57 1.43 GHz & 4Gb LPDDR4 1600MHz 64 bit\\
      \hline
  
    \end{tabular}
  \end{center}
\end{table}

\noindent\textbf{Software}\\
To provide developers with a stable versioned codebase to work with, Open Source Robotics Foundation (OSRF) releases a set of different ROS packages known as distribution \cite{Ros2Distribution}. ROS distributions directly depend on the Linux distribution. According to the table \ref{tab:hardware}, we have different physical devices with different CPUs, such as ARM-64 and X64, so the Operating systems must also be different to work on them. We chose Ubuntu 20.04 LTS AMD 64  \cite{Ubuntupc} for laptop computers and Ubuntu 20.04 LTS ARM 64  \cite{Ubuntuarm} for the Pi 3, Pi 4, and Jetson Nano (4GB) sd card image \cite{Jetbotsdcard} for Jetson Nano. For our evaluation, we used ROS2 Humble \cite{Humble},  the latest ROS2 distribution. ROS2 Humble can be installed only on Ubuntu 22.04, and different DDS implementations have different pre-requisite versions of libraries. To solve the previously mentioned problems, we have used docker to keep an abstraction between the base operating system, pre-requisite libraries, and DDS implementations. Moreover, docker produces negligible latency overhead while running on the host network \cite{DockerVsVM}, so it simplifies our hardware-software building blocks and does not affect the evaluation measurement. ROS2 provides two client libraries, rclcpp (for C++) and rclpy (for Python), to create nodes, subscribers, and publishers. We have used the rclpy client library to write and build our ROS2 nodes, topics, data transfer, and communications. \\

\begin{table}[h!]
  \begin{center}
    \caption{Variable data type, size, and frequency .}
    \label{tab:datatype}
    \begin{tabular}{|c|p{2in}|p{2in}|} 
    \hline
      \textbf{Data Type} & \textbf{Data Size} & \textbf{Publisher Frequency Hz} \\
      \hline
        Binary & 33Kb, 67Kb, 87Kb, 145Kb, 294Kb, 502Kb, 1Mb, 2Mb, 4Mb   &  1 2 4 8 10 12 15 20 100 1000   \\
     \hline
      String &  0.004KB, 0.091KB, 1.405KB, 2.675KB &  10 20 40 80 100 200 500 1000\\
      \hline
       IMU &  & 10 20 40 80 100 200 500 1000\\
      \hline

    \end{tabular}
  \end{center}
\end{table}

\begin{figure}[h!]
\centering
  \includegraphics[width=\textwidth]{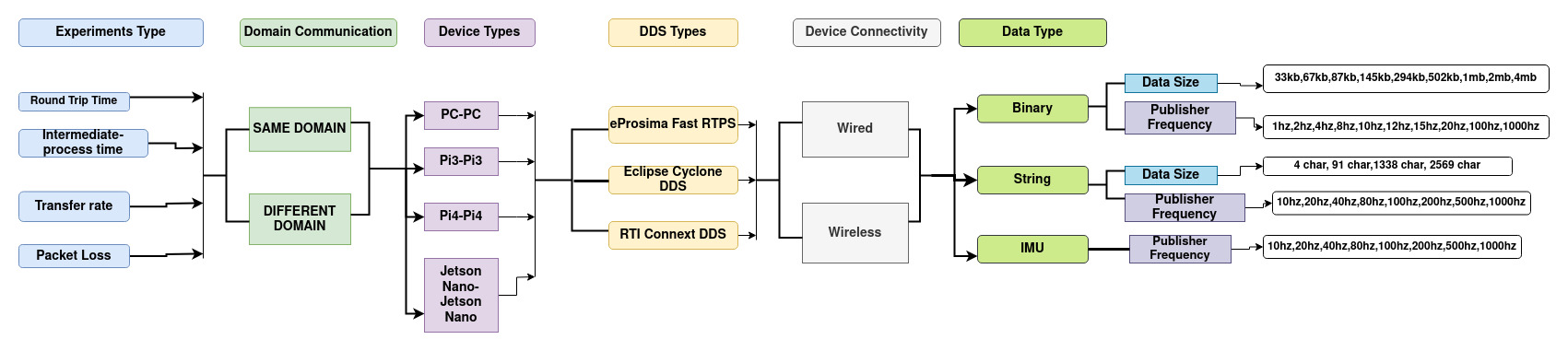}
  \caption{ Experiment scenarios.}
  \label{fig:experimentflow}
\end{figure}

\noindent\textbf{Experiment Scenario}\\
The possible number of experiments is huge because we have different numbers of parameters subsection-\ref{parameter} for the evaluation process. Figure \ref{fig:experimentflow} represents all the desired experiments as a flow chart from left to right.

\section{Evaluation}
\begin{figure}[h]
    \centering
    \begin{subfigure}[b]{0.48\textwidth}
        \centering
        \includegraphics[width=\textwidth]{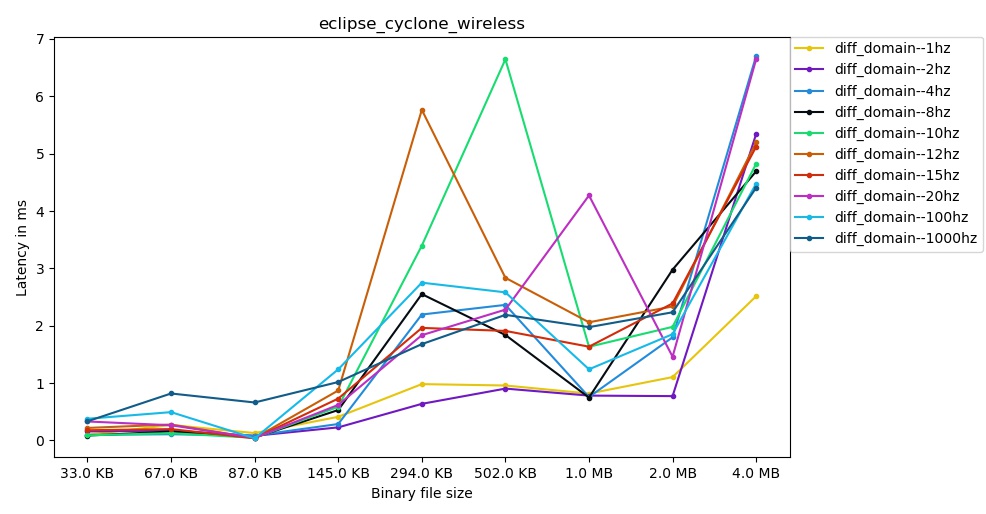} 
        \caption{Eclipse Cyclone Laptop DD-Wireless}
        \label{sub:cyclone_pc_wireless_dd}
    \end{subfigure}
    \begin{subfigure}[b]{0.48\textwidth}
        \centering
        \includegraphics[width=\textwidth]{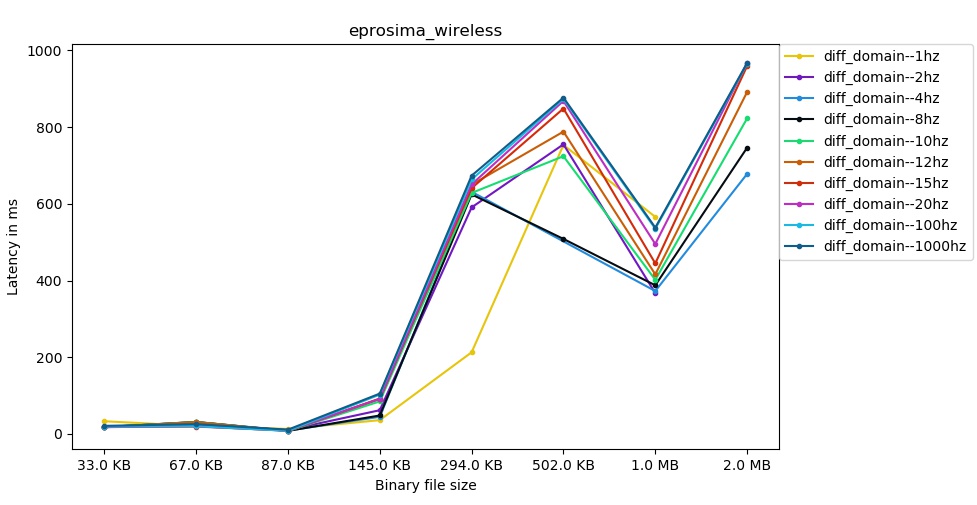} 
        \caption{eProsima Fast RTPS Pi4 DD-Wireless}
        \label{sub:eprosima_pi4_wireless_dd}
    \end{subfigure}
    
    \vskip\baselineskip
    
    \begin{subfigure}[b]{0.48\textwidth}
        \centering
        \includegraphics[width=\textwidth]{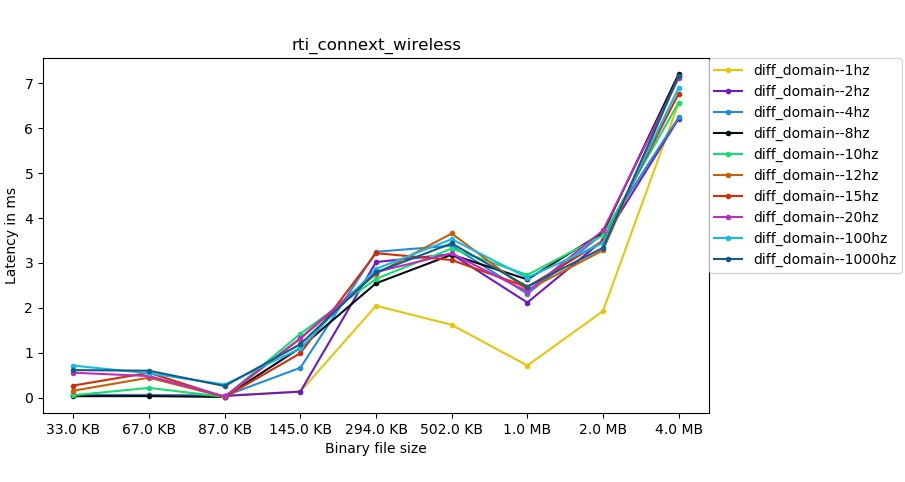} 
        \caption{RTI Connext Laptop DD-wireless}
        \label{sub:rti_pc_wireless_dd}
    \end{subfigure}
    \begin{subfigure}[b]{0.48\textwidth}
        \centering
        \includegraphics[width=\textwidth]{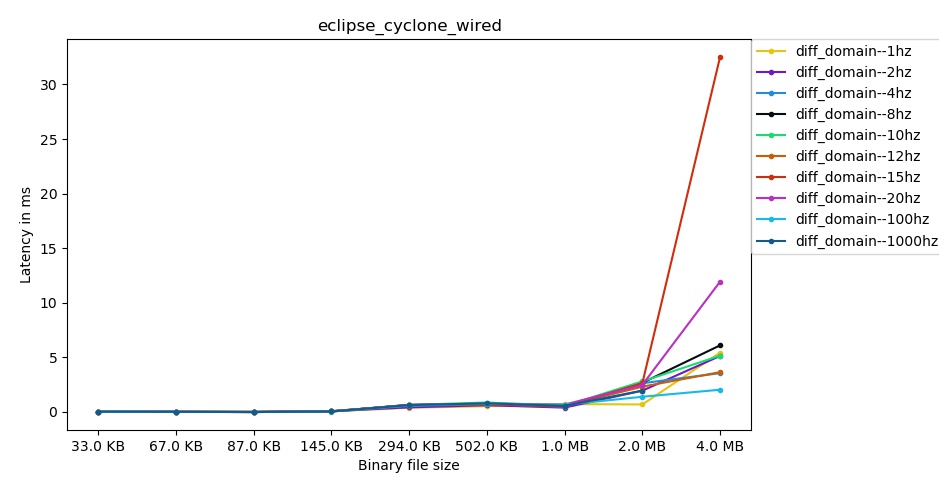} 
        \caption{Eclipse Cyclone DD-Wired}
        \label{sub:cyclone_pi3_wired_dd}
    \end{subfigure}
    \caption{For the Different Domain or \(DD\) wireless connectivity, all vendor-specific DDS implementations have sudden spikes in latency when file size increases than 145KB, but for wired connection, it shows consistent latency for Eclipse Cyclone DDS.}
    \label{fig:main}
\end{figure}

\noindent\textbf{Effect of Frequency on Latency}\\
In our research, we observed intriguing patterns in communication latency across different domains and connection types for Eclipse Cyclone DDS, eProsima RTPS and RTI Connext DDS. We found that when communicating across different domains, varying frequencies led to different latency patterns for different file sizes. Notably, for Eclipse Cyclone DDS, higher frequencies resulted in increased latency for larger files, as illustrated in figure-\ref{sub:cyclone_pc_wireless_dd}. We were particularly interested to see a sudden spike in latency for both wirelessly connected nodes (same and different domain) when the file size reached 502KB. Interestingly, we observed higher latency for a 502KB file size compared to a 1MB file size for both eProsima Fast RTPS and RTI Connext DDS figure-\ref{sub:rti_pc_wireless_dd}, \ref{sub:eprosima_pi4_wireless_dd}. However, our different domain communication experiment for wired connectivity for Eclipse Cyclone DDS revealed consistent latency characteristics across all file sizes (figure-\ref{sub:cyclone_pi3_wired_dd}), which contrasts with the behavior we saw in wirelessly connected different domain communication.

\begin{figure}[h!]  
\begin{subfigure}{.48\textwidth}
\centering
  \includegraphics[width=\textwidth]{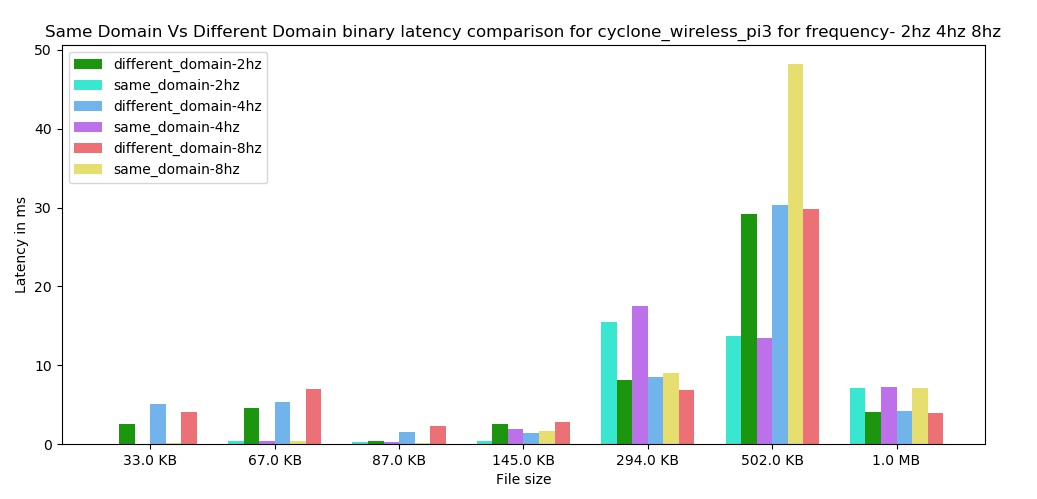}
  \caption{}
  \label{sd_vs_dd_cyclone_pi3_wireless_low}
\end{subfigure}
\begin{subfigure}{.48\textwidth}
\centering
  \includegraphics[width=\textwidth]{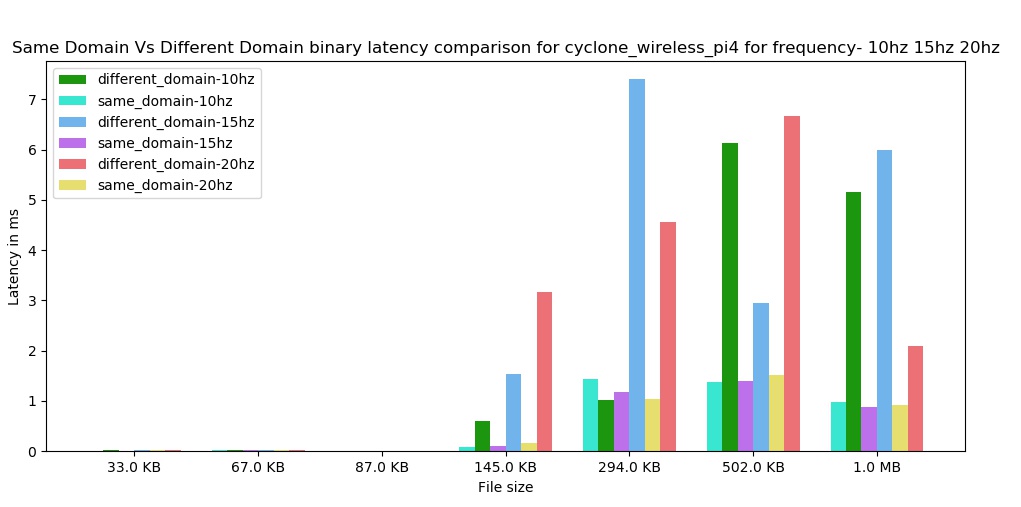}
  \caption{}
  \label{sd_vs_dd_cyclone_pi4_wireless_high}
\end{subfigure}
\caption{Eclipse Cyclone DDS same domain vs. different domain for binary data wireless connectivity Pi3 Pi4 node.}
\label{fig:sd_vs_dd_cyclone_wireless_pi3}
\end{figure}

\noindent\textbf{Same Domain vs Different Domain Communication}\\
\noindent\textbf{Binary Data}\\
In our experiments with ROS2 nodes using different DDS implementations, we observed various interesting patterns in latency characteristics across same and different domain communications. For Pi3 nodes using Cyclone DDS with wireless connections, we found that same domain communication outperformed different domain communication for smaller file sizes figure-\ref{sd_vs_dd_cyclone_pi3_wireless_low}. However, this trend reversed for larger files, particularly at 1MB, where different domain communication showed better performance across all low publisher frequencies. Our Pi4 node experiment revealed different behavior figure-\ref{sd_vs_dd_cyclone_pi4_wireless_high}, with same domain communication generally performing better as file sizes increased.

\begin{figure}[h]  
\begin{subfigure}{.48\textwidth}
\centering
  \includegraphics[width=\textwidth]{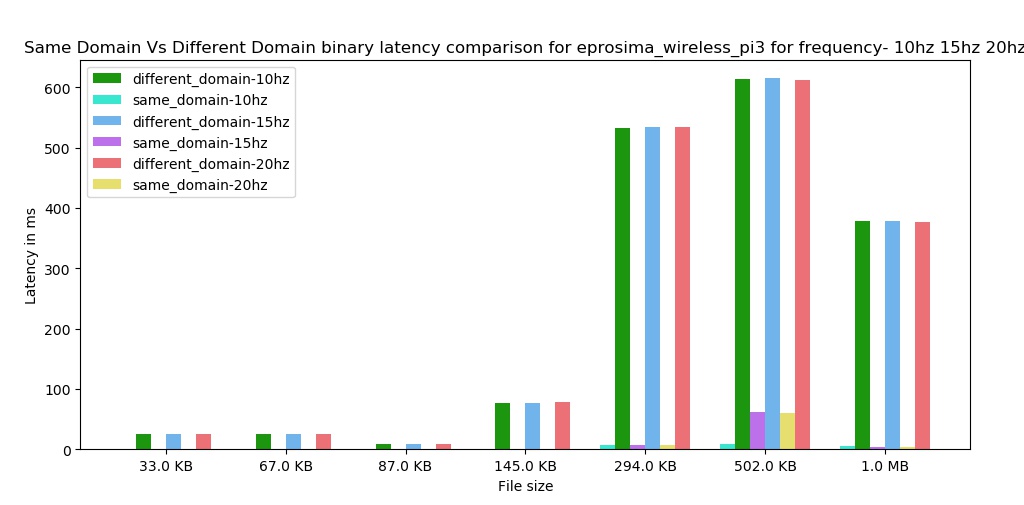}
  \caption{}
  \label{sd_vs_dd_eprosima_pi3_wireless_low}
\end{subfigure}
\begin{subfigure}{.48\textwidth}
\centering
  \includegraphics[width=\textwidth]{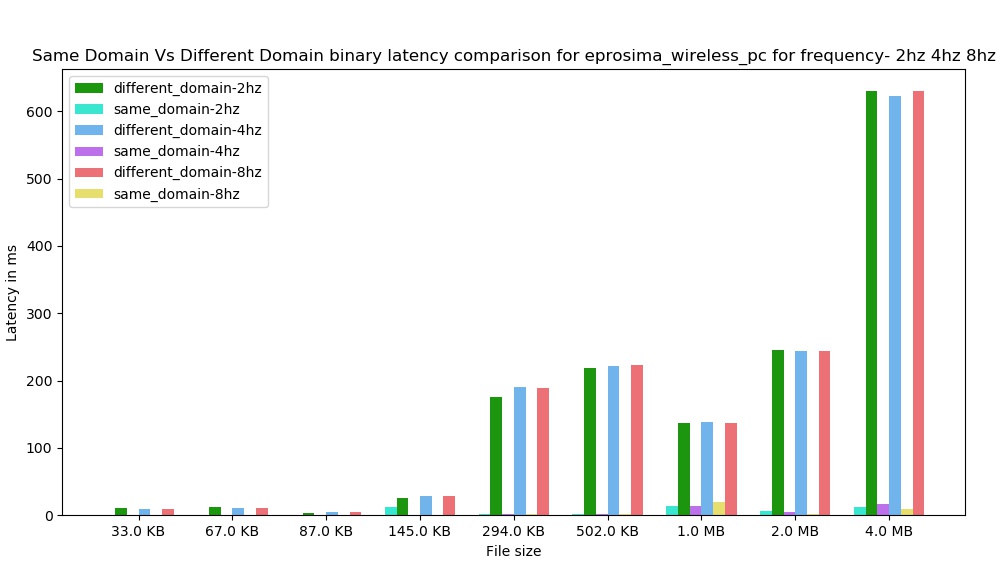}
  \caption{}
  \label{sd_vs_dd_eprosima_pi3_wired_high}
\end{subfigure}
\caption{eProsima DDS same domain vs. different domain for binary data}
\label{fig:sd_vs_dd_eprosima_pi3}
\end{figure}

Interestingly, when we switched to wired connections, the latency characteristics changed significantly for both Pi3 and Pi4, with same domain communication showing lower latency for larger files on Pi3 - a complete reversal from the wireless scenario.\\
When using eProsima DDS, we observed consistent behavior across all machine types and connectivity options. Same domain communication consistently outperformed different domain communication for all file sizes and publisher frequencies figure-\ref{fig:sd_vs_dd_eprosima_pi3}.

\begin{figure}[h!]  
\begin{subfigure}{.48\textwidth}
\centering
  \includegraphics[width=\textwidth]{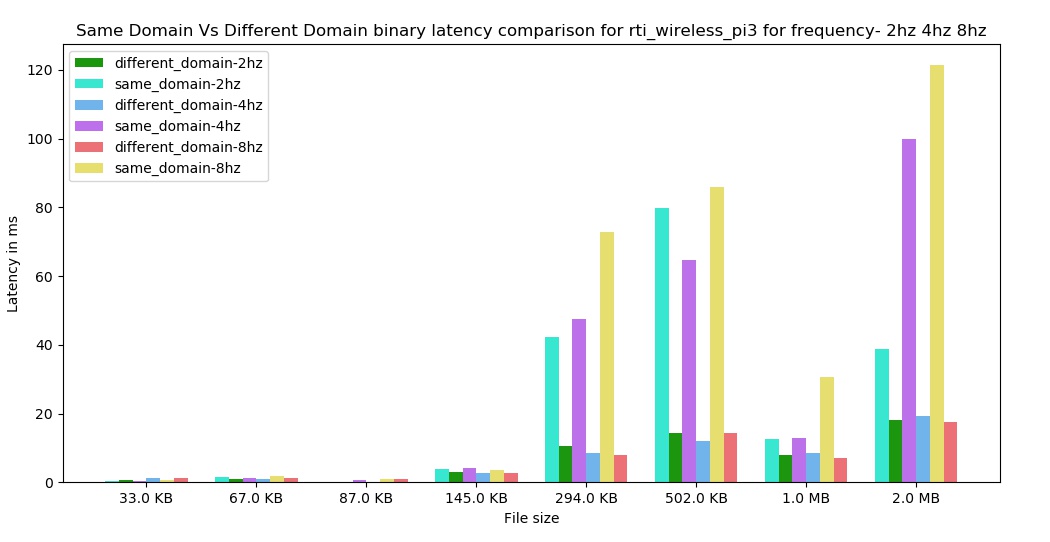}
  \caption{}
  \label{sd_vs_dd_rti_pi4_wireless_low}
\end{subfigure}
\begin{subfigure}{.48\textwidth}
\centering
  \includegraphics[width=\textwidth]{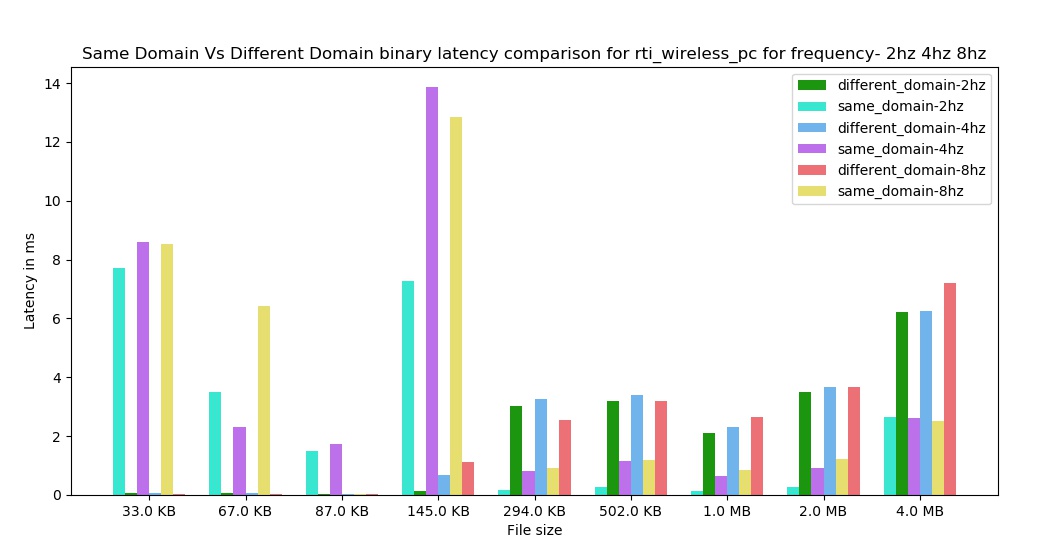}
  \caption{}
  \label{sd_vs_dd_rti_pc_wireless_low}
\end{subfigure}
\caption{RTI Connext DDS same domain vs. different domain wireless connectivity (a)Pi3 (b)Laptop node.}
\label{fig:sd_vs_dd_rti_wireless_pi4}
\end{figure}

RTI Connext DDS showed more varied behavior. For wirelessly connected Pi3 and Pi4 nodes at lower frequencies, we saw similar latency for both domains up to 87KB file size. However, for Pi3, different domain communication performed better for files larger than 145KB figure-\ref{sd_vs_dd_rti_pi4_wireless_low}. Laptop nodes exhibited yet another pattern figure-\ref{sd_vs_dd_rti_pc_wireless_low}, with same domain communication showing higher latency for smaller files up to 145KB when connected wirelessly.

In wired connections using RTI Connext DDS, we observed minimal latency differences between domains for files up to 145KB across all node types. For larger files on Pi3, same domain communication showed lower latency. Interestingly, wired laptop nodes displayed different behavior compared to their wireless counterparts, with different domain communication showing significantly lower latency for most file sizes, except for 2MB files where the difference was minimal.

\noindent\textbf{String and IMU data}\\
For String and IMU data, Pi3, Pi4, and laptop computer wired and wireless connectivity, same-domain and different-domain communication nearly performs similarly for Eclipse Cyclone DDS and RTI Connext DDS. On the other hand, eProsima Fast RTPS different-domain communication performs worst, like Binary data transfer.\\
\\
 We noticed that publisher frequency significantly impacted latency in most cases, except for wirelessly connected Raspberry Pi3 nodes using eProsima DDS, where latency remained consistent across frequencies. Unexpectedly, 502KB binary files often yielded higher latency than 1MB files across all DDS implementations. We also observed that for smaller file sizes up to 145KB, eProsima DDS, Eclipse Cyclone DDS, and sometimes RTI Connext DDS showed similar latency characteristics regardless of publisher frequencies.
When comparing same-domain and different-domain communication, we found varying results across DDS systems, file types, and sizes. For instance, with Eclipse Cyclone DDS, same-domain communication generally performed better for wired connections, while different-domain communication excelled in wireless setups for larger files. eProsima DDS consistently showed better performance in same-domain communication across all scenarios. RTI Connext DDS exhibited more complex behavior, with performance varying based on node type, connection method, and file size.
For string and IMU data exchange, we generally observed better performance in same-domain communication across all DDS implementations. However, in our Realsense live camera stream experiments with Jetson Nano nodes, RTI Connext DDS showed superior performance in different-domain communication.

Based on the previous discussion, it is not advisable to use eProsima DDS for communication across different domains. ROS2 developers can achieve better performance when transferring large files among wirelessly connected different domains by using Eclipse Cyclone DDS. RTI Connext DDS shows a smaller difference in performance between same domain and different domain communication. Therefore, if a ROS2 developer requires both types of communication, it is recommended to use RTI Connext DDS as the rmw.


\section{Future Work}
In future work, we aim to expand our research in ROS2 communication and cooperative driving systems. We plan to explore a broader range of QoS policy settings and their compatibility, conduct node stress testing to evaluate performance under system and network load and perform mobile node experiments to simulate real vehicle scenarios. We also intend to integrate 5G technology to investigate its impact on communication performance, analyze the effects of implementing ROS2 security packages, and develop a GUI-based framework that leverages machine learning to recommend optimal DDS implementations and settings based on user requirements. We are interested in examining the impact of security on the performance of different domain communication. We want to investigate why 502KB data frames are experiencing a sudden latency increase. Through this comprehensive approach, we hope to address critical challenges in autonomous vehicle communication, enhance ROS2 usability, and contribute significantly to the optimization of cooperative driving systems. Our goal is to provide valuable insights and tools that will advance the field of autonomous and cooperative driving.

\section{Conclusion}

In the autonomous and cooperative perception paradigm, transferring hundreds of sensor data among different connected vehicles using only a single ROS2 domain with a limited number of participants and topics is not realistic. Different domain communication must be used to communicate with a higher number of participants and fulfill the requirement of a flexible number of topics. This thesis paper presents the first performance evaluation between ROS2 nodes situated in different domains and communicating via wire and wireless connectivities. The evaluation involved different physical machines and file types. From the evaluation, we observed that no single vendor-specific DDS implementation dominates during all use case experiments.

In most cases, the same domain communication outperformed different domain communication with good margins. However, there are some instances where different domain communication is more efficient for transferring larger files, resulting in more packets being transferred. eProsima DDS performed worst in all use cases regarding different domain communication for wired and wireless connectivity for all data types. This study provides guidance for robot system developers and researchers on choosing the appropriate data type, minimum data size, connectivity medium, and vendor-specific DDS implementation to ensure optimal communication among ROS2 nodes. If we consider different domain ROS2 communication over wireless medium as the base communication method for the autonomous vehicles and cooperative driving paradigm, from our experiments, we can say that at this moment, ROS2's vendor-specific DDS implementations are not capable enough to fulfill the hard real-time requirements. The possible reason might be the involvement and performance of bridging services between different domain communication. In order to facilitate communication among an unlimited number of ROS2 nodes or autonomous vehicles and ensure real-time end-to-end latencies for real-time systems, it is necessary for all vendors to collaborate on joint research to develop new mechanisms and improve different domain communication.

\bibliography{references}

\appendix

\section{Online Resources}

The Source code for the experiment can be found here.
\begin{itemize}
\item \href{https://github.com/sumitpaulde/ros2-dds-performance-evaluation}{GitHub- https://github.com/sumitpaulde/ros2-dds-performance-evaluation},

\end{itemize}

\end{document}